\pdfoutput=1

\documentclass[11pt]{article}

\usepackage{emnlp2021}

\usepackage{times}
\usepackage{latexsym}

\usepackage[T1]{fontenc}

\usepackage[utf8]{inputenc}

\usepackage{booktabs}

\usepackage{titlesec}

\usepackage{microtype}

\usepackage{graphicx}
\usepackage{amsmath}
\usepackage{amssymb}
\usepackage{subfigure} 
\usepackage{color}
\usepackage{numprint}
\usepackage{multirow,array,varwidth}
\usepackage{booktabs}
\usepackage{makecell}
\usepackage{bbm}
\usepackage{dsfont}
\usepackage{xspace}
\usepackage{amsthm}
\usepackage{float}

\makeatletter
\g@addto@macro\small{%
  \setlength\abovedisplayskip{-5pt}
  \setlength\abovedisplayshortskip{-5pt}
  \setlength\belowdisplayshortskip{-5pt}
  \setlength\belowdisplayskip{-5pt}
}
\makeatother

\newcommand{\emoji}[1]{\includegraphics[height=.025\textwidth]{#1}}

\newcommand{\train}{{\footnotesize\texttt{train}\xspace}}
\newcommand{\val}{{\footnotesize\texttt{val}\xspace}}
\newcommand{\test}{{\footnotesize\texttt{test}\xspace}}

\newcommand{\longan}{Alternating Normalization\xspace}
\newcommand{\longcan}{Classification with Alternating Normalization\xspace}
\newcommand{\longanlc}{alternating normalization\xspace}

\newcommand{\an}{\textsc{AN}\xspace}
\newcommand{\can}{\textsc{\footnotesize{CAN}}\xspace}
\newcommand{\baseline}{\textsc{\footnotesize{Baseline}}\xspace}
\newcommand{\bprediction}{source prediction}
\newcommand{\rset}{reference set}

\let\vec\mathbf 
\newcommand{\R}[0]{\mathds{R}}
\newcommand{\eqcomma}[0]{\;\;,}
\newcommand{\eqdot}[0]{\;\;.}

\theoremstyle{definition}
\newtheorem{definition}{Definition}

\DeclareMathOperator*{\argmax}{arg\,max}

\newcommand{\diag}{\mathcal{D}}

\makeatletter
\DeclareRobustCommand\onedot{\futurelet\@let@token\@onedot}
\def\@onedot{\ifx\@let@token.\else.\null\fi\xspace}

\def\ie{\emph{i.e}\onedot}

\def\wrt{w.r.t\onedot}

\makeatother

\newcommand{\meanstd}[2]{#1}

\definecolor{green_im}{rgb}{0.1, 0.55, 0.3}

\newcommand{\Rise}[1]{\textcolor{green_im}{\scriptsize{(\bf $\uparrow$#1})\xspace}}

\newcommand{\appxsec}[1]{Appendix~\ref{#1}}

\newcommand{\eat}[1]{}

\title{When in Doubt: {I}mproving Classification Performance with\\ Alternating Normalization}

\author{
  Menglin Jia$^{1,2}$\hspace{8pt}
  Austin Reiter$^{2}$\hspace{8pt}
  Ser-Nam Lim$^{2}$\hspace{8pt}
  Yoav Artzi$^{1}$\hspace{8pt}
  Claire Cardie$^{1}$\\
$^{1}$Cornell University\qquad $^{2}$Facebook AI 
\\\texttt{\{mj493,yya5,claire\}@cornell.edu,\{areiter,sernamlim\}@fb.com}
} 

\date{}

\begin{document}
\maketitle

\begin{abstract}
We introduce \longcan (\can), a non-parametric post-processing step for classification. 
\can improves classification accuracy for challenging examples by re-adjusting their predicted class probability distribution using the predicted class distributions of high-confidence validation examples. 
\can is easily applicable to any probabilistic classifier, with minimal computation overhead. We analyze the properties of \can using simulated experiments, and empirically demonstrate its effectiveness across a diverse set of classification tasks~\footnote{Our code for the work are open
sourced at \href{https://github.com/kmnp/can}{\texttt{github.com/kmnp/can}}. }.

\end{abstract}

\section{Introduction}\label{sec:intro}

Classification is core to NLP, and many language problems can be effectively addressed as supervised classification tasks. 
However, even the most effective classifier can suffer when given examples to classify that are close to its decision boundary. The reasons for such failures vary, and include lack of training data coverage, limited representation expressivity, or over-fitting the training data. Despite significant progress, including using pre-trained models~\cite{devlin-etal-2019-bert} to address these issues, every classifier has its weak spots, and some examples will be hard to classify correctly. 

In this paper, we study a simple non-parameterized post-processing step to improve classifier accuracy on difficult examples. 
At the core of our approach is using \emph{\longan}~\cite[\an;][]{sinkhorn1967concerning} to re-adjust the prediction of low-confidence examples using the predicted class distributions of a \emph{\rset{}} of high-confidence validation examples. 

\begin{figure}[t]
\centering
\includegraphics[width=\columnwidth]{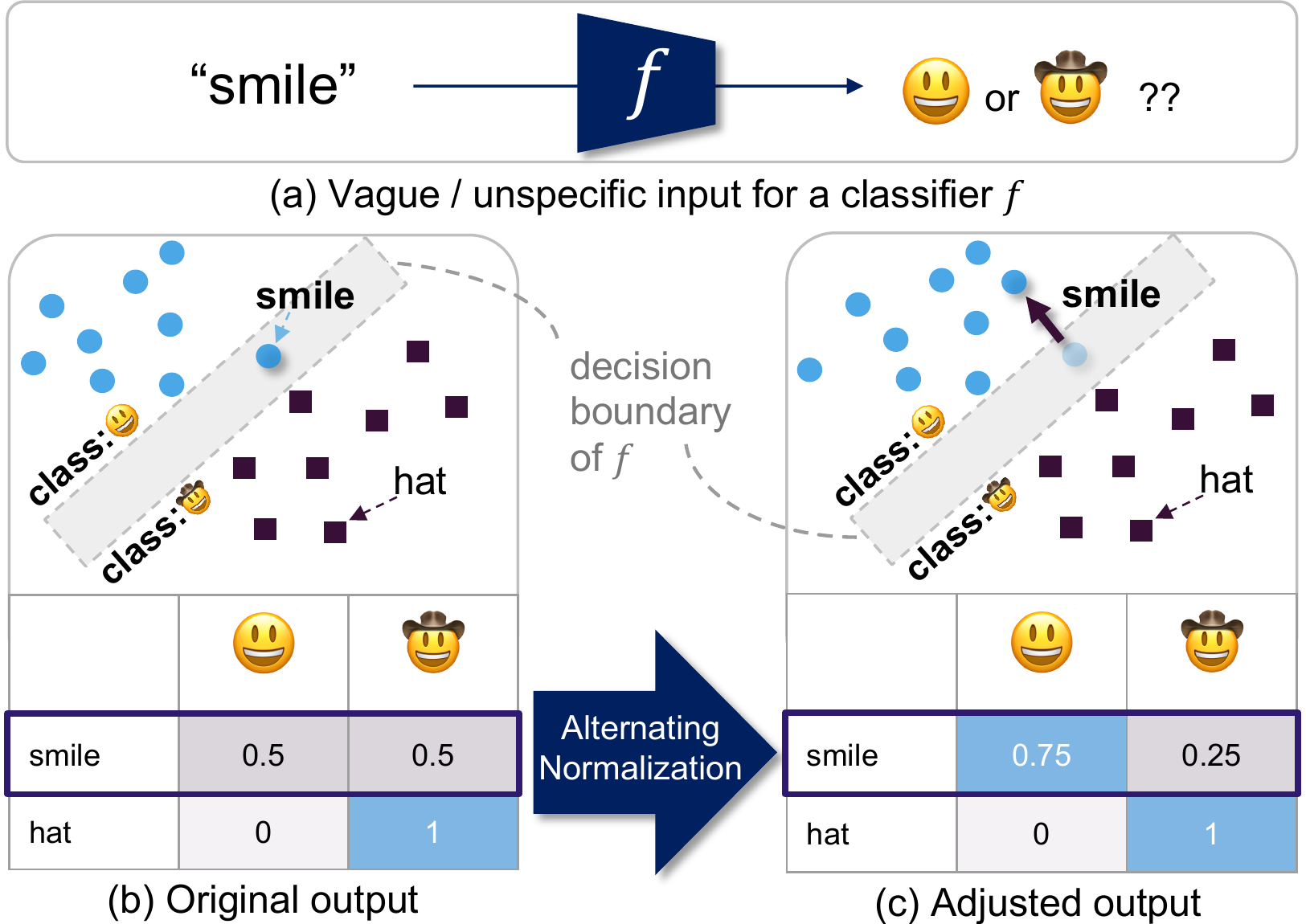}
\caption{An illustration of \longcan (\can).
Given an example ``smile'', the classifier $f$ assigns equal probability to both emojis (b). \can re-scales the class probability distribution and produces a less ambiguous prediction (c).\footnotemark[1]
}
\label{fig:teaser}
\end{figure}

Our process, \longcan (\can), is applicable to any classifier that generates a distribution over target classes.
We first identify challenging examples and a disambiguating \rset{} using the ambiguity level of the predicted class probability distributions.
Then we perform a series of normalizations, alternating between normalizing across examples for each class and for each example across classes. 
For example, in Figure~\ref{fig:teaser}, we classify an input example (``smile'') to one of two labels (\emoji{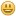} and~\emoji{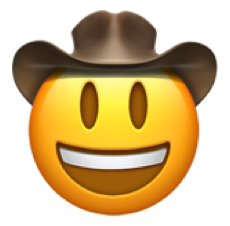}). The example sits at the decision boundary between the two target labels, and is completely ambiguous. The reference set includes a single example (``hat''), which the classifier can resolve with high confidence. Here, we use a single alternating normalization step, including normalizing across rows (examples) and columns (target labels), which disambiguates this simple example to classify it correctly.

\footnotetext[1]{We use a classical pragmatic reasoning example for our illustration, highlighting our inspiration in the Rational Speech Act~\cite[RSA;][]{frank2012predicting} model, which we discuss in Section~\ref{sec:rsa}.}\stepcounter{footnote}

We study \can on randomly generated matrices (Section~\ref{sec:sim}) and evaluate it on several text classification tasks (Section~\ref{sec:exp}). 
In general, we find that \can is most effective when the original predictions are of higher ambiguity. Our experiments also suggest \can is more beneficial for tasks with many labels. 
 
\section{\longcan}\label{sec:method}

We modify the output of a probabilistic classifier for an ambiguous input example at test time using \longanlc  with respect to an additional set of examples to re-adjust the input example's predicted probabilities.
Let $\mathcal{X}$ and $\mathcal{Y}$ be random variables for input and labels, respectively.
Given a \emph{challenging} example $x \in \mathcal{X}$, and a probability distribution $\mathbb{P}(\mathcal{Y} = y\vert \mathcal{X} = x)$ over a set of $m$ classes produced by a probabilistic classifier, we compute $\mathbb{P}^{\prime}(y \vert x)$ as an adjusted distribution in three steps, the second of which is an iterative process.

Our method requires a \rset{} of examples. 
We use the confident portion of the validation data commonly used in standard evaluation settings. 
Given $n$ confident examples, we create a row stochastic matrix $A_0 \in \R^{n \times m}$ by concatenating the predicted probability distributions.
We describe the proposed method below.

\paragraph{Step 1: Identify Hard Examples}

Let $\vec{b}_0 \in \R^m$ be the \bprediction{}, \ie, the predicted class probability distribution for a challenging example $x$.
We identify examples that can benefit from \can 
by computing the \textit{ambiguity level} of its class probability distribution $\vec{b}_0$.
With a selected example, we construct a matrix  $L_0 \in \R^{(n + 1) \times m}$ by concatenating $A_0$ and $\vec{b}_0$ along the rows:\footnote{$\vec{b}_0^T$ denotes the transpose of $\vec{b}_0$.} 

\begin{small}
\begin{equation}
    L_0 = \begin{bmatrix} A_0 \\ \vec{b}_0^T \end{bmatrix}\;\;.
\end{equation}
\end{small}

A common way to identify challenging examples is by measuring the entropy of their predicted class distribution $\mathcal{H}(\vec{b}_{0}) = -\sum_i \vec{b}_{0,i} \log_m (\vec{b}_{0,i})$.
The higher the value of $\mathcal{H}(\vec{b}_0)$ is, the more uniform the distribution $\vec{v}$ is, which indicates higher ambiguity level.
We observe $\mathcal{H}(\vec{b}_0)$ may not be ideal to capture ambiguity well for our classification purpose, which is concerned mainly with high probability events. 
For example, consider two distributions $\vec{b}_0^1 = \begin{bmatrix}0.5 &0.25 &0.25 \end{bmatrix}$ and $\vec{b}_0^2 = \begin{bmatrix}0.5 &0.5 &0 \end{bmatrix}$, for which $\mathcal{H}(\vec{b}_0^1) > \mathcal{H}(\vec{b}_0^2)$. 
However, $\vec{b}_0^2$ expresses a more uncertain classification result.

Instead, we select challenging examples based on the  entropy near the peak of the distribution.
We define top-$k$-entropy to focus on the top of the distribution. 
Let $\mathcal{T}: \R^m \rightarrow \R^{k}$ be the ${\rm top}$-$k$ operator. 
The top-$k$-entropy is:

\begin{small}
\begin{equation}
\label{eq:topk_entropy}
    \mathcal{H}_{\rm top\text{-}k}(\vec{b}_0) = \mathcal{H}(\mathcal{T} (\vec{b}_0, k) )\eqdot
\end{equation}
\end{small}

\noindent
We use a base of $m$ so that  $0 < \mathcal{H}_{\rm top\text{-}k}(\vec{b}_0) \leq 1 $.

Given a scalar threshold $0\leq \tau \leq 1$ and the number of classes $k_{max}$,
the ambiguity level of a probability distribution $\vec{b}_0$ is larger than $\tau$, if for any $k \in  [2, k_{max}]$, the top-$k$-entropy of $\vec{b}_0$ is greater than $\tau$.

\paragraph{Step 2: One Iteration of \an} 

Each iteration $d$ of \an normalizes $L_0$, first across its rows (row norm) and then along its columns (column norm).  Let $\diag: \R^n \rightarrow \R^{n\times n}$  turn a vector $\vec{v}$ into a diagonal matrix, and let $\vec{e}$ be a vector of ones\footnote{\appxsec{subsec:geo} provide a step-by-step visualization of the \longanlc{} process, and \appxsec{subsec:converge} discusses the its convergence}.

\subparagraph{Step 2.1: Row Norm}
The \emph{row normalization} of $L_{d-1}$ is: 

\begin{small}
\begin{align}
    \Lambda_S &= \diag(\left( L_{d-1}^{\alpha}\right)^T \vec{e}) \label{eq:Lambda_S}\\
    S_d &= L_{d-1}^{\alpha} \Lambda_S^{-1}\eqcomma
    \label{eq:row_norm}
\end{align}
\end{small}

\noindent
where $\alpha > 0$, $L_{d-1}^{\alpha}$ is the matrix exponentiation of $L_{d-1}$, and $\Lambda_S^{-1}$ is the inverse of $\Lambda_S$.
The diagonal entries of $\Lambda_S \in \R^{m \times m}$ represent the column sums of $L_{d-1}^{\alpha}$, so that $S_d$ is column stochastic. 
The parameter $\alpha$ controls the rate of convergence of $\vec{b}_0$  to a high confidence state.

\subparagraph{Step 2.2: Column Norm}
The \emph{column normalization} step is:

\begin{small}
\begin{align}
    \Lambda_L &= \diag(S_d \Lambda_q \vec{e}) \\
    L_{d} &= \Lambda_L^{-1} S_d \Lambda_q\eqcomma \label{eq:col_norm}
\end{align}
\end{small}

\noindent
where $\Lambda_q \in\R^{m \times m}$ is a diagonal matrix that represents the class priors, which we approximate using the training class distribution.
The diagonal entries of $\Lambda_L\in \R^{(n + 1) \times (n + 1)}$ are the row sums of $S_d \Lambda_q$ so that $L_d$ is row stochastic.\footnote{Each normalization step (Steps 2.1--2.2) takes $\mathcal{O}(mn)$ because the matrices $\Lambda_S, \Lambda_L, \Lambda_q$ are diagonal.}

\paragraph{Step 3: Re-adjusted Output Extraction}
Let $L_d$ be the resulting matrix after $d$ steps of Step~2:

\begin{small}
\begin{equation}
\label{eq: Ld_matrix}
    L_d =  \begin{bmatrix} A_d \\ \vec{b}_d^T \end{bmatrix}\eqdot
\end{equation}
\end{small}

\noindent
We keep $\vec{b}_d$ as the re-adjusted class probability distribution $\mathds{P}^{\prime}(y \vert x)$, and discard $A_d$.


\begin{table*}[t]
\small
\begin{center}
\resizebox{\textwidth}{!}{%
\begin{tabular}{l l l   r r    r r}
\toprule
\multirow{2}{*}{\textbf{Datasets}} 
&  \multirow{2}{*}{\textbf{$\#$ Classes}}
&  \multirow{2}{*}{\textbf{Method}}
&\multicolumn{2}{ c }{\textbf{Marco F1}}
&\multicolumn{2}{ c }{\textbf{Mirco F1}}
\\
\cmidrule{4-7}
&&&\val{} & \test{}  &\val{} & \test{} \\ 
\midrule

\multirow{4}{*}{\shortstack[l]{Ultrafine\\Entity\\Typing}}
&\multirow{4}{*}{10331}
&\baseline (Multitask;~\citealp{choi-etal-2018-ultra})
&31.32	 & 31.98 &27.92 &28.80  \\
&&\can{}
&\Rise{2.15} \textbf{33.47} &\Rise{1.71} \textbf{33.69}
&\Rise{2.51} \textbf{30.43} &\Rise{1.89} \textbf{30.69}  \\
\cmidrule{3-7}

&&\baseline (Denoised;~\citealp{onoe-durrett-2019-learning})
& 40.07	  & 40.22 & 37.88  & 37.87  \\
&&\can{}
&\Rise{0.34} \textbf{40.41} &\Rise{0.53} \textbf{40.75}
&\Rise{0.59} \textbf{38.47} &\Rise{0.84} \textbf{38.71}  \\
\midrule

\multirow{4}{*}{DialogRE}
&\multirow{4}{*}{36}
&\baseline (BERT;~\citealp{yu-etal-2020-dialogue} )
&\meanstd{35.89}{0.81} &\meanstd{35.76}{1.70} 
&\meanstd{59.44}{0.61} &\meanstd{57.93}{0.89} \\
&&\can{}	
&\Rise{0.91} \meanstd{\textbf{36.80}}{1.19}	
&\Rise{0.70} \meanstd{\textbf{36.45}}{1.88}
&\Rise{0.16} \meanstd{\textbf{59.60}}{0.60}	
&\Rise{0.34} \meanstd{\textbf{58.27}}{1.01} \\
\cmidrule{3-7}

&&\baseline (BERTs;~\citealp{yu-etal-2020-dialogue})
&\meanstd{40.58}{0.54} &\meanstd{39.45}{1.92}	
&\meanstd{62.18}{1.14}	&\meanstd{59.52}{1.90}	\\
&&\can{}
&\Rise{0.83} \meanstd{\textbf{41.41}}{0.95}	
&\Rise{0.68} \meanstd{\textbf{40.13}}{2.00}	 
&\Rise{0.33} \meanstd{\textbf{62.51}}{1.11}	
&\Rise{0.29} \meanstd{\textbf{59.81}}{1.77} \\
\bottomrule

\end{tabular}
}
\caption{Performance on the Ultrafine Entity Typing and DialogRE tasks.}
\label{tab:all-results}
\end{center}
\end{table*}

\begin{figure}
\centering
\includegraphics[width=\columnwidth]{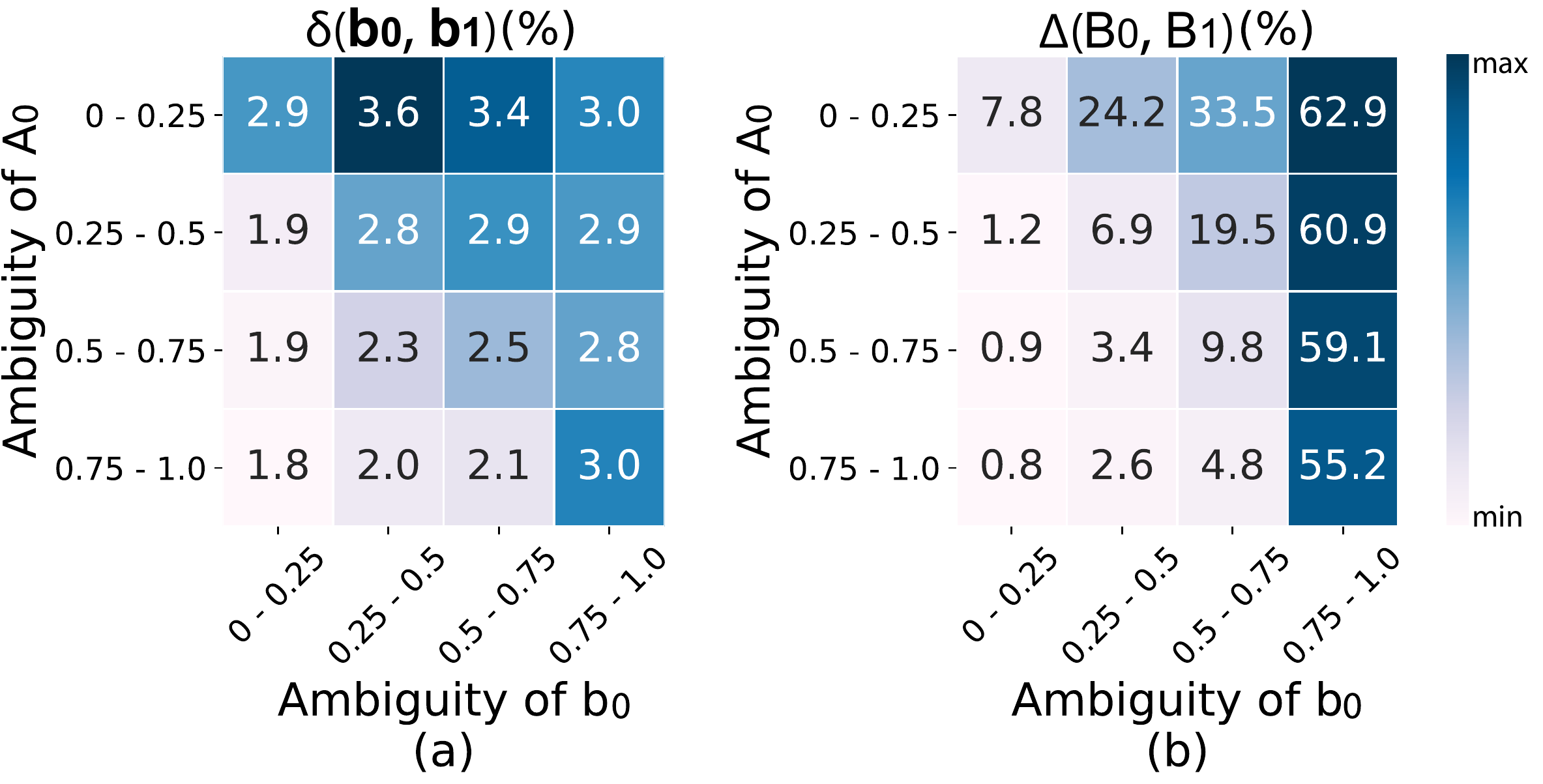}
\caption{Averaged expected accuracy gain ($\delta(\vec{b}_0, \vec{b}_1)$) and classification accuracy gain ($\Delta(B_0, B_1)$). Original prediction $\vec{b}_0$ with high ambiguity level yields higher expected accuracy gain.
}
\label{fig:sim_all}
\end{figure}

\section{Simulations on Random Matrices}\label{sec:sim}

We study the effect of the ambiguity level of the \bprediction{} and the \rset{} using Monte Carlo simulations. We randomly generate $A_0$, $\vec{b}_0$, and $\Lambda_q$ to evaluate the expected performance change after each iteration of \an ($\vec{b}_0 \rightarrow \vec{b}_1$) as a function of the ambiguity level of $\vec{b}_0$ and $A_0$.

\paragraph{Setup}
The ambiguity levels are grouped into 4 intervals: $\{[0, 0.25)$, $[0.25, 0.5)$, $[0.5, 0.75)$, $[0.75, 1]\}$.
Given the number of classes $m$ and an ambiguity interval, we randomly generate $A_0\in \R^{(m-1) \times m}$,$\Lambda_q$, and $B_0\in \R^{n \times m}$ independently,\footnote{Similar to~\citet{yuan2018understanding}, we set $L$ / $S$ as square matrices for simplicity. In practice, $L$ / $S$ do not need to be square, as shown in Section~\ref{sec:exp}.}
where each row of $B_0$ represents a randomly generated $\vec{b}_0$.
For each interval and each $m \in \{2, \dots ,10, 20, \dots, 100\}$, we randomly generate $A_0$, $B_0$, and $\Lambda_q$ 200 times with $n = 100$.
We compute each $\vec{b}_1$ separately using \can with $\alpha \in \{0.1, \dots 0.9, 1, \dots 9\}$, and construct $B_1$.

\paragraph{Evaluation Metrics}
We define two metrics to evaluate the expected classifier performance change: (a) 
$\delta(\vec{b}_0, \vec{b}_1)$ measures the expected performance gain of $\vec{b}_1$ \wrt $\vec{b}_0$; and (b) $\Delta(B_0, B_1)$ measures the accuracy gain (i.e., of the $\arg\max$) of a set of input examples $B_1$ \wrt $B_0$.\footnote{\appxsec{sec:more_sim} provides formal definitions.}
We explore how $\delta(\vec{b}_0, \vec{b}_1)$ and $\Delta(B_0, B_1)$ change as a function of the ambiguity level of $B_0$ and $A_0$.

\paragraph{Effect of Ambiguity Level}
Figure~\ref{fig:sim_all} shows the averaged $\delta(\vec{b}_0, \vec{b}_1)$ and $\Delta(B_0, B_1)$ across all matrix sizes, simulations, and values of $\alpha$.
We observe that (a) the expected accuracy gains are positive across all ambiguity levels of $A_0$ and $\vec{b}_0$; (b) \can{} tends to improve the performance of $\vec{b}_0$ with high ambiguity level, especially using a \rset{} with low top-$k$ entropy; and (c) the performance is robust to the ambiguity level of $A_0$.

\begin{figure*}[t]
\centering
\includegraphics[width=\textwidth]{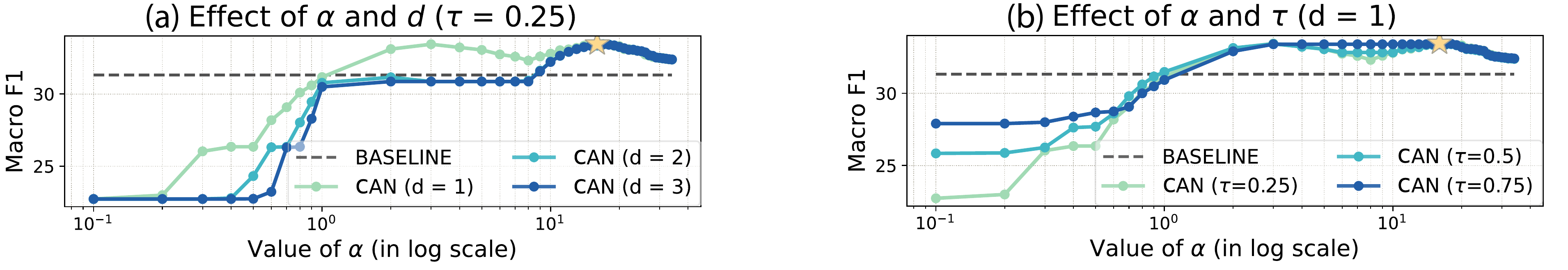}
\caption{Effect of $\alpha$, recursion depth $d$, and $\tau$ on the macro F1 scores of ultrafine entity typing \val{} set with the \baseline multitask classifier.}
\label{fig:hp_tune}
\end{figure*}

\section{Empirical Experiments}\label{sec:exp}

We evaluate \can{} using three classification tasks: ultrafine entity typing~\cite{choi-etal-2018-ultra}, and dialogue-based relation extraction~\cite[DialogRE;][]{yu-etal-2020-dialogue}. 
We compare off-the-shelf classifiers (\baseline) and our method.\footnote{\appxsec{subsec:expsetup} provides details for \baseline{} methods.}
We select \can's hyperparameters ($\alpha$, $d$, $\tau$) using the official validation sets (\val{}) and evaluate on the official test sets (\text{}). 
The challenging subsets of the \val{} and \test{} are identified as the \bprediction{}s using top-$k$ entropy. The rest of \val{}, which has low ambiguity, is used as the \rset{} $A_0$.\footnote{\appxsec{suppsec:imp} and~\ref{supsec:detail} provide implementation and reproducibility details.}

\paragraph{Results} 

Table~\ref{tab:all-results} summarizes the experimental results.
\can{} offers consistent performance gains for different classifiers and datasets by re-adjusting the uncertain examples only, especially when the task has many classes.
Table~\ref{tab:all-results} suggests more effective classifiers benefit less from \can, but still see improvements. 
For example, in the ultrafine entity typing experiment, we observe a larger improvement for
the multitask version~\cite{choi-etal-2018-ultra} compared with the denoised one~\cite{onoe-durrett-2019-learning}, which uses a cleaned up version of the training data.

\eat{For models with multiple runs for DialogeRE and MDID, we also report the performance gain on each run in Figure~\ref{fig:per_run}. 
This further demonstrate the generalization ability of the proposed method \can{}.}

\paragraph{Tuning \can{}} Figure~\ref{fig:hp_tune} shows the how the values of the hyperparameters ($\alpha, d, \tau$) affect \can{} on the \val{} set using ultrafine entity typing multitask \baseline{}.
We observe that the effect of $d$ and $\tau$ diminish gradually as $\alpha$ grows, because $\alpha$ controls how quickly \can{} transforms the \bprediction{}s to high-confidence ones. 
For a fixed recursion depth $d$, the performance does not always improve using a larger $\alpha$.
This suggests that larger $\alpha$ can deteriorate the performance through over-calibration.
We also see that a small value of recursion depth $d$ yields the best results in general.

\section{Analysis}\label{subsec:ana_impact}

We hypothesize that classifiers with better performance require less re-adjustment. 
We test this hypothesis by controlling the number of ``hard'' examples, therefore controlling the performance of the \baseline{} classifier. We use ImageNet~\cite{imagenet_cvpr09} because of the existence of established ways for image perturbation without modifying the image semantics. 
This dataset contains 1.3 million training and 50k validation natural scene images with 1000 classes.
We use a ResNet-50~\cite{he2016deep} model from the default pretrained models in torchvision~\cite{paszke2017pytorch} as \baseline{}.

\begin{figure}
\centering
\includegraphics[width=0.9\columnwidth]{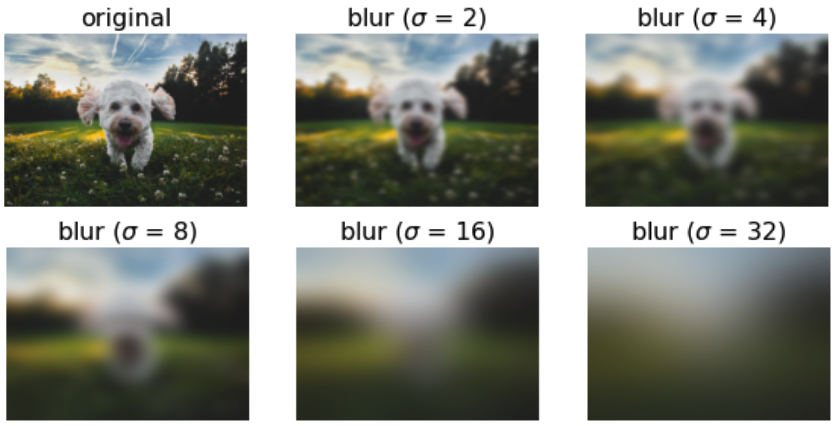}
\caption{Image blurring examples.}
\label{fig:blur}
\vspace{+5pt}
\end{figure}

We systematically make the task harder by convolving the images with a Gaussian function with zero-mean and various values of standard deviation $\sigma$. 
Higher values of $\sigma$ results in more blurred image, which emulate examples with higher uncertainty and lower \baseline{} classifier performance. 
Figure~\ref{fig:blur} shows example images with different values of $\sigma \in \{2, 4, 8, 16, 32\}$.

We assess the effect of \can by comparing two variants of \can with \baseline{} using this setup:
(a) \can{}\textsc{-naive}: $\alpha = 1.0$, recursion depth is 1;
and (b) \can{}\textsc{-best}: an upperbound-version of \can with $\alpha$ and depth $d$ optimized on the test data.

\begin{figure}
\centering
\includegraphics[width=0.95\columnwidth]{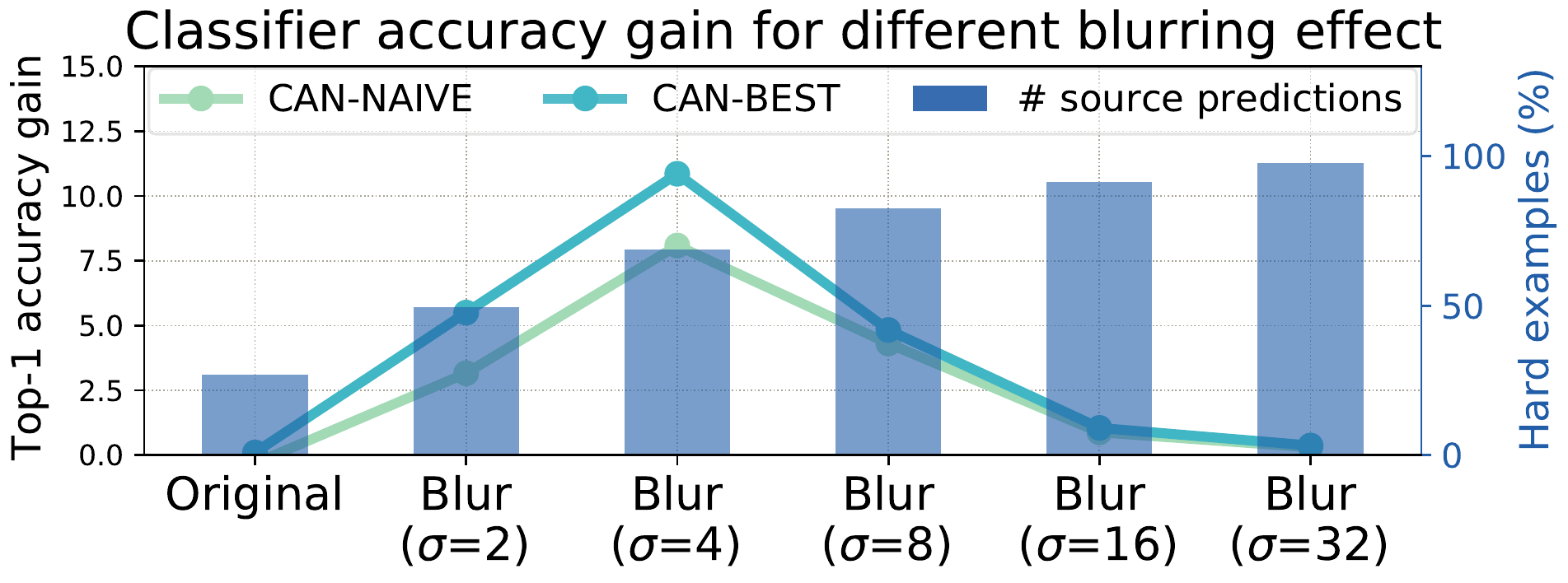}
\caption{Effect on accuracy of making examples ``harder'' by blurring. }
\label{fig:blur_results}
\end{figure}

Figure~\ref{fig:blur_results} presents the absolute top-1 accuracy ($\%$) gain on the hard subset of \val{}. 
Optimizing $\alpha$ and $d$ (\can{}\textsc{-best}) indeed offers larger performance gains across six different image settings comparing with \can{}\textsc{-naive}.
We observe a general positive correlation between the number of hard examples and the relative gain using \can{}, confirming that our method benefits classifiers that receive harder data. 
However, this advantage diminishes when a classifier significantly under-performs, as the accuracy gain dropped for Blur ($\sigma$ = 8) onward and its top-1  accuracy is only 3.25$\%$ on hard subset.

\section{Related Work}\label{sec:rsa}

\paragraph{Rational Speech Act (RSA)} 
RSA is a framework for pragmatic reasoning, where a speaker and a listener generate and understand utterances by reasoning about the understanding and intentions of their interaction partner~\cite{frank2012predicting,goodman2016pragmatic}. 
Both agents are probabilistic, and RSA uses alternating normalization in a recursive process. 
Our technique is motivated by the type of reasoning provided by RSA, whereas interpretation and generation of messages are considered within the context of other interpretations and messages to resolve ambiguities. 

We adapt the RSA technique to post-process the output of probabilistic classifiers. 
The classifier takes a similar role to the RSA listener in the alternating normalization process. 
The matrices $S_d$ and $L_d$ from Equations~\ref{eq:row_norm} and~\ref{eq:col_norm} are aligned with the stochastic matrices for the derived speaker and listener in RSA.
The row norm (Equation~\ref{eq:row_norm}) mirrors the special case of a speaker that considers the cost of generating a message as zero. 
This assumption is reasonable for classification. It is also common practice when working with a finite set of intents in RSA~\cite{monroe-etal-2017-colors, zarriess-schlangen-2019-know}. 
The column norm (Equation~\ref{eq:col_norm}) describes a mathematical formulation aligned with how a pragmatic listener infers speaker expectations.

Previous work has applied RSA to systems that generate and understand language~\citep{andreas-klein-2016-reasoning, mao2016generation, Vedantam_2017_CVPR, cohn-gordon-etal-2018-pragmatically, zarriess-schlangen-2019-know} in both referential games~\cite{frank2012predicting, goodman2016pragmatic, monroe-etal-2017-colors} and sequential decision-making systems~\cite{fried-etal-2018-unified,NEURIPS2018_6a81681a}.
Our method departs from these applications by 
focusing on the ambiguity avoidance property of the listener agent as applied to generic classification tasks.

\paragraph{Confidence Calibration}
Similar to confidence calibration techniques~\citep{platt1999probabilistic, zadrozny2002transforming, guo2017calibration, kumar2019calibration}, our method rescales the posterior distribution produced by the classifier at test time. However, the aim of calibration is to make  the output probabilities more representative of the correctness likelihood, whereas our's is to resolve ambiguity.

\section{Conclusions and Future Work}
We propose \longcan{} (\can) as a simple and light-weight post-processing step. 
Our method adjusts the predicted class distribution of ``edge cases'' for a generic classifier during test time.
Via experiments on both simulated and real-world NLP tasks,
we show that \can{} helps improve performance using a fixed \rset{} with low ambiguity and increases the performance of standard classifiers.
Future work may further investigate  the properties of \can. For example, \appxsec{subsec:ana_impact} describes an initial study on vision examples showing the benefit of \can increases as examples become noisier, even with the same classifier. 
Advancing this study and generalizing it to language is an important direction for future work. 
Finally, one could study improving \can, for example by selecting $A_0$ strategically, or applying it during training.

\section*{Acknowledgements}
We would like to thank the members of the Cornell NLP group, and anonymous reviewers for their helpful feedback.
Photo of Figure~\ref{fig:blur} by~\href{https://unsplash.com/@joeyc}{Joe Caione} on~\href{https://unsplash.com/@joeyc?utm_source=unsplash&utm_medium=referral&utm_content=creditCopyText}{Unsplash}.
\bibliography{anthology,custom}
\bibliographystyle{acl_natbib}

\clearpage
\newpage
\appendix

\section{Implementation Details}
\label{suppsec:imp}

\subsection{Extension to Multilabel Problems}
The problem we describe in Section~\ref{sec:method} is supervised multi-class classification.
For multi-label classification task, given $n$ input examples and $m$ classes, a classifier usually produces an array of size $n \times m$ where the value in each position represents the predicted probability of one input-class pair.
We transform this array into shape $nm \times 2$, where each row is the binary probability distribution for each input-class pair.
This is similar to the common binary cross-entropy loss for this task.

\subsection{Simulation Metrics Definitions} \label{sec:more_sim}

Under the framework described in Section~\ref{sec:method}, we define two metrics to evaluate the expected classifier performance.
\emph{Expected performance gain} measures the performance of $\vec{b}_1$ \wrt $\vec{b}_0$.
Let $\mathcal{Y}$, $\hat{\mathcal{Y}}$ be two random variables. $\mathcal{Y}$ represents the true label of a given instance, and $\hat{\mathcal{Y}}$ is the predicted class label out of a set of classes $C$.
The \emph{expected accuracy} of $\vec{b}_0$ would be:

\begin{small}
\begin{align}
    \mathbb{E}_0[Acc] &= \frac{1}{\vert C \vert} 
   \sum_{c \in C} \mathbb{P}(\mathcal{Y}=c) \mathbb{P}(\hat{\mathcal{Y}}=c)  \\
   &=  \frac{1}{\vert C \vert} \vec{q}^T \vec{b}_0\eqcomma
\end{align}
\end{small}

\noindent
where each entry at position $c$ of $\vec{q} \in \R^{\vert C \vert}$ is the probability mass function (PMF) of $\mathcal{Y}$ when $\mathcal{Y} = c$. $\vec{q}$ is the main diagonals of the randomly generated $\Lambda_q$.
Each entry at position $c$ of $\vec{b}_0$ is the PMF of $\hat{\mathcal{Y}}$ when $\hat{\mathcal{Y}} = c$. 
The relative expected performance gain is then defined as:

\theoremstyle{definition}
\begin{definition}[Expected Performance Gain]
\label{def:Delta}
The relative \emph{expected performance gain} of $\vec{b}_1$ \wrt $\vec{b}_0$ is:

\begin{small}
\begin{align}
    \delta (\vec{b}_0, \vec{b}_1)
    =\frac{ \mathbb{E}_1[Acc] - \mathbb{E}_0[Acc] } {\mathbb{E}_0[Acc] }\eqdot
\end{align}
\end{small}
\end{definition}

\noindent
$\delta$ quantifies the performance on individual example. Next we introduce a second metric, \emph{accuracy gain}, to measure a set of input examples. 

\theoremstyle{definition}
\begin{definition}[Accuracy Gain] 
\label{def:flip_accu}
Given $B_1 \in \R^{n \times m} $, where the $i^{th}$ row represent a re-adjusted predicted distribution, denoted as $\vec{b}_1^{i}$, the overall performance of $B_1$ \wrt $B_0$ is evaluated as:

\begin{small}
\begin{multline}
\Delta(B_0, B_1) = \\
 \frac{1}{n}\sum_i \mathds{1}\{ \delta_i >0, \argmax_{\vec{b}_0^{i}} \neq \argmax_{\vec{b}_1^{i}}\}\eqcomma
\end{multline}
\end{small}

\noindent
where $\delta_i = \delta (\vec{b}_0^{i}, \vec{b}_1^{i})$.
We consider \can{} is successful when the predicted classes change from $\vec{b}_1$ to $\vec{b}_0$ and $\delta (\vec{b}_0, \vec{b}_1) > 0 $.  $\Delta$ is used for the primary metric in our simulation study.
\end{definition}




\section{Technical Analysis}
\label{suppsec:ana}

\subsection{Geometric Interpretation}
\label{subsec:geo}
\begin{figure}[t]
\centering
\subfigure[$L_{d-1}$.]{
    \includegraphics[scale=0.365]{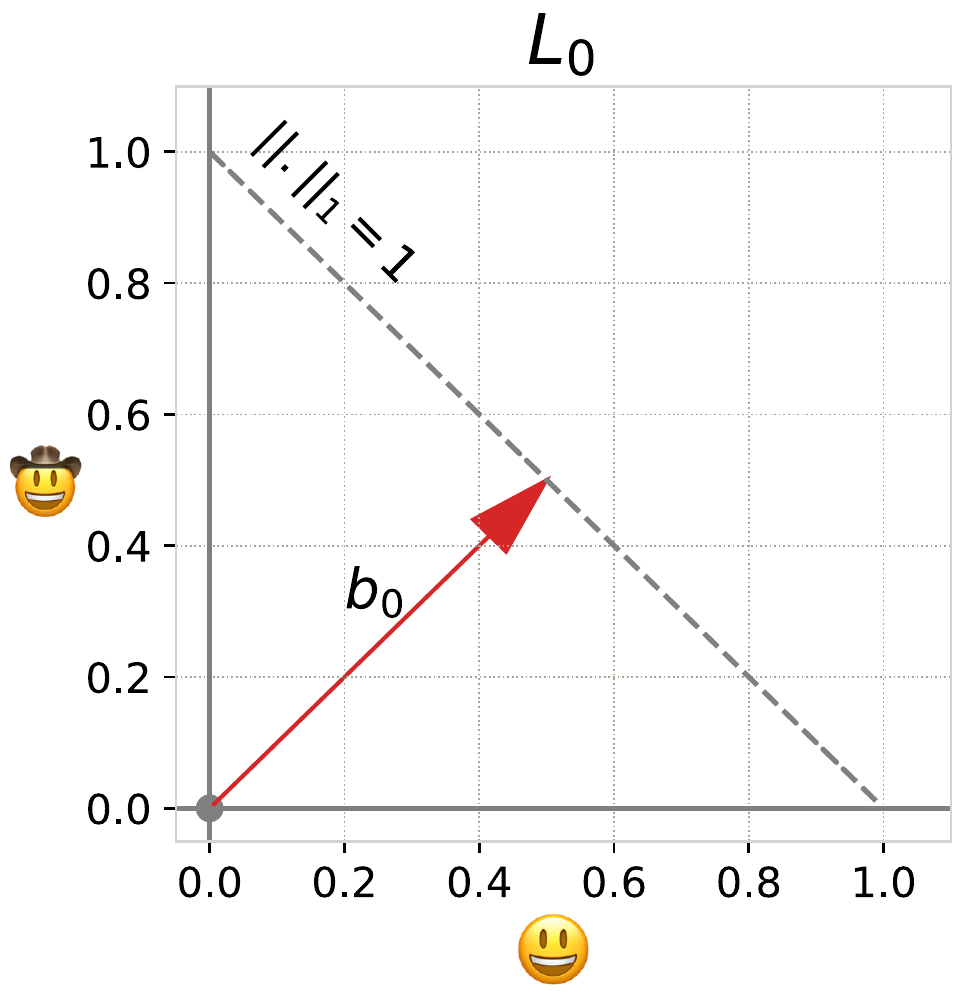}
    \label{fig:trans_ld-1}
}
\hfill
\subfigure[$S_d$: effect of $\Lambda_1$.]{
    \includegraphics[scale=0.365]{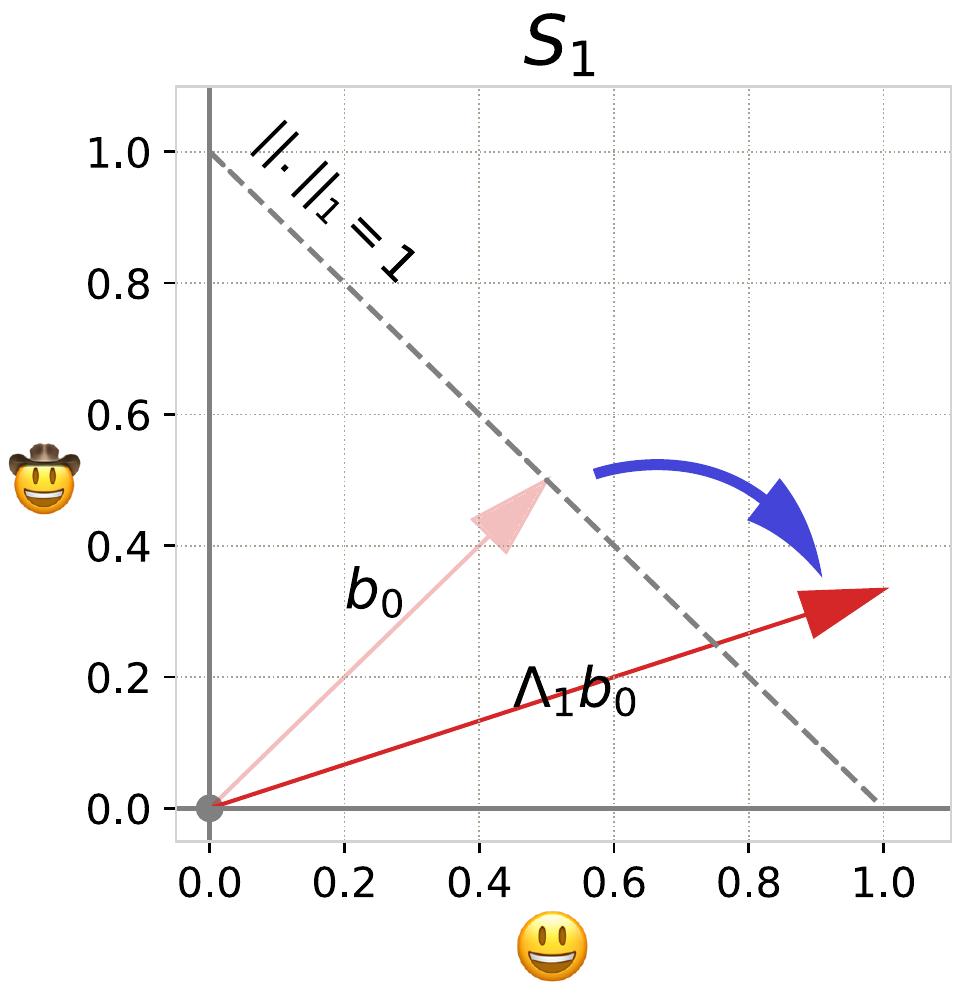}
    \label{fig:trans_sd}
}
\hfill
\subfigure[$L_d$: effect of $\lambda_2$.]{
    \includegraphics[scale=0.365]{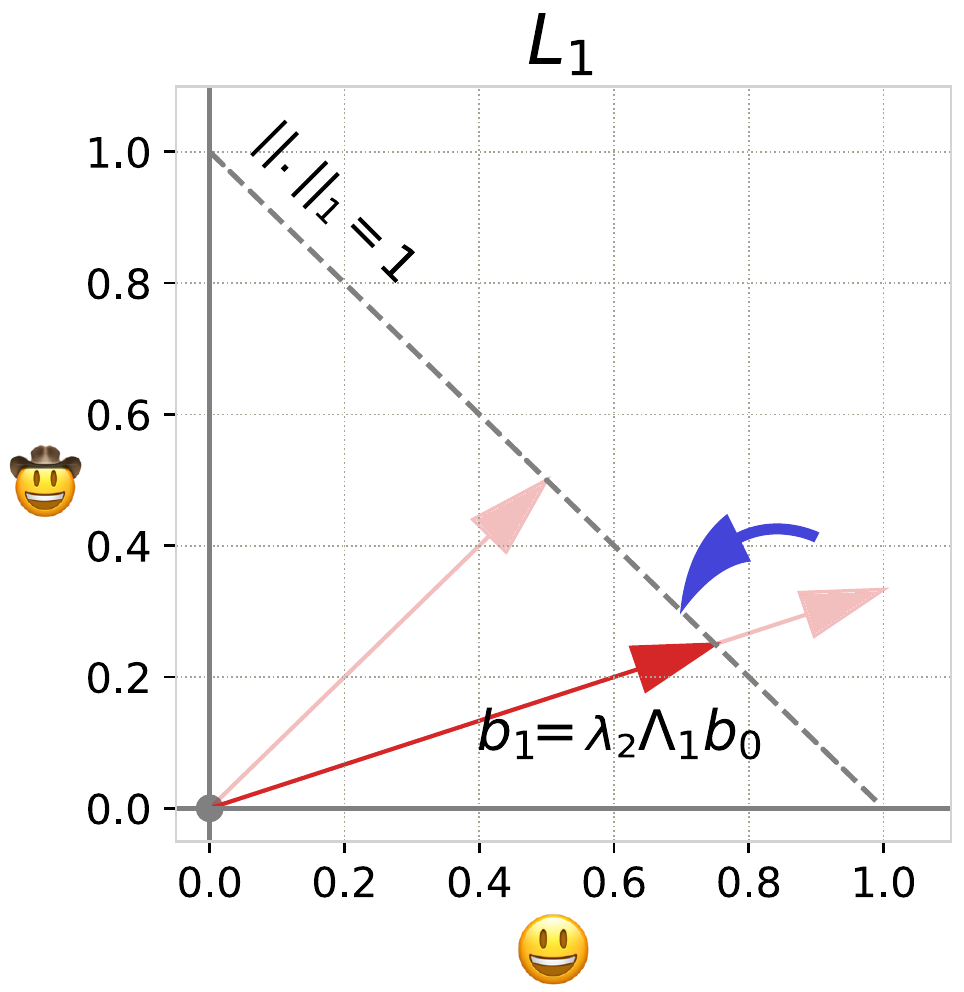}
    \label{fig:trans_ld}
}
\hfill
\subfigure[Effect of $\alpha$.]{
    \includegraphics[scale=0.365]{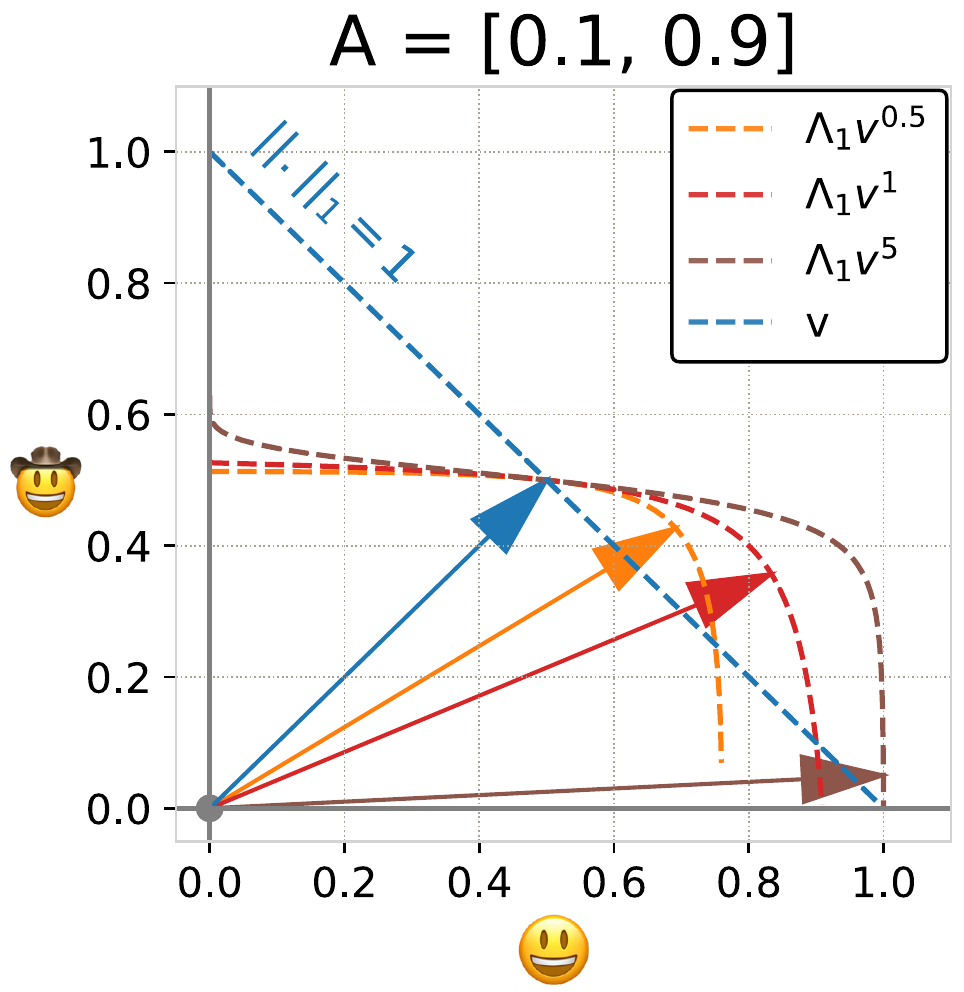}
    \label{fig:eigen_91}
}
\caption{Visualization of $\vec{b}_0 \rightarrow \vec{b}_1$ in the row space of $L$.
Dashed straight line represents the unit ball of $\R^2$ with respect to $\|\cdot \|_1$ in first quadrant. All transformations assume a uniform prior.
Figure~\ref{fig:trans_sd}~\ref{fig:trans_ld} plot Row Norm and Column Norm respectively, given $A_0 = \begin{bmatrix}
0 &1\end{bmatrix}$ and $\alpha=1$.
Figure~\ref{fig:eigen_91} shows the effect of $\alpha$ using a non-unit $A$ for better visual effect.
}
\label{fig:an_vis}
\end{figure}

To intuitively understand the~\longanlc,
we visualize the effect of the Row Norm and Column Norm in Section~\ref{sec:method} to the \bprediction{} in Figure~\ref{fig:an_vis}. It portrays a step-by-step transformations of the \bprediction{} $\vec{b}_0 \rightarrow \vec{b}_1$, where $\vec{b}_0, \vec{b}_1 \in \R^2$ using the example in Figure~\ref{fig:teaser}.

Consider two diagonal matrices $\Lambda_1, \Lambda_2$, where $\Lambda_2 = \Lambda_L^{-1} \in \R^{(n+1) \times (n+1)}$, $\Lambda_1 = \Lambda_S^{-1}\Lambda_q\in \R^{m \times m}$.
$\Lambda_L, \Lambda_S, \Lambda_q$ are the scaling matrices in Eq.~\ref{eq:Lambda_S}-\ref{eq:row_norm}.
The transformation of $L_{d-1} \rightarrow L_d$ can be written as:

\begin{small}
\begin{equation}
\label{eq:L_trans}
    L_{d} = \Lambda_2 L_{d-1}^{\alpha} \Lambda_1\eqdot
\end{equation}
\end{small}

\noindent
Since we care about the transformation of $\vec{b}_d$, combining Eq.~\ref{eq: Ld_matrix} and~\ref{eq:L_trans} yields the main operations of our method:

\begin{small}
\begin{equation}
\label{eq:b_trans}
    \vec{b}_d = \lambda_2 \Lambda_1 \vec{b}_{d-1}^{\alpha}\;\;,
\end{equation}
\end{small}

\noindent
$\lambda_2$ is a scalar at the $(n+1, n+1)$ entry of $\Lambda_2$.

\paragraph{Visualizing the effect of $\Lambda_1,\lambda_2$}
Figure~\ref{fig:trans_ld-1}-\ref{fig:trans_ld} show that
starting from $\vec{b}_0$, which has unit 1-norm, $\Lambda_1$ rotates and scales $\vec{b}_0$ (Figure~\ref{fig:trans_sd}), $\lambda_2$ scales the resulting vector so $\vec{b}_1$ has unit 1-norm (Figure~\ref{fig:trans_ld}).
$\Lambda_1$ determines which direction $\vec{b}_0$ rotates to. 
Intuitively speaking, it is the context-aware nature of the RSA that governs the transformation of $\vec{b}_0$.

\paragraph{Effect of $\alpha$}
Figure~\ref{fig:eigen_91} shows the trajectory of $\Lambda_1 \vec{v}^{\alpha}$ for $\vec{v}$ that has unit 1-norm in a toy example where $L \in \R^{2 \times 2}$, $A_0 = \begin{bmatrix}0.1 &0.9\end{bmatrix}$.
The larger the value of $\alpha$, the closer $\Lambda_1 \vec{v}^{\alpha}$ is to the base vector in the row space of $L$. This observation reconfirms the effect of $\alpha$ in Section~\ref{sec:exp}: it controls how much to take into account of the input vector $\vec{v}$ and how quickly for $\vec{v}$ to achieve a high confident state.

\subsection{Convergence of $L_d$}
\label{subsec:converge}
Let $\alpha=1$ and $\Lambda_q$ be an identity matrix.
Given a square $L_d$, \citet{sinkhorn1967concerning} have proved that in this setting, $\lim_{d\to\infty} L_d$ will converge if $L_0$ is full rank.
Consider a $2\times2$ case, where $A_0 = \begin{bmatrix}p_A & 1-p_A\end{bmatrix}$, $\vec{b}_0^T = \begin{bmatrix}p & 1 - p \end{bmatrix}$.
According to the Sinkhorn-Knopp algorithm, if $p_A(1-p) - p(1 - p_A) \neq 0$, 

\begin{small}
\begin{equation}
    \begin{bmatrix}p_A & 1-p_A  \\p & 1 - p \end{bmatrix}
    \xrightarrow{d \rightarrow \infty} 
    \begin{bmatrix}1 & 0 \\ 0 & 1\end{bmatrix} \text{or} \begin{bmatrix}0 & 1 \\ 1 & 0\end{bmatrix}\eqdot
\end{equation}\end{small}

\noindent
Therefore, $\lim_{d\to \infty} \vec{b}_d$ will always converge to a unit vector which has the lowest entropy possible if $p_A \neq p$.

This illustrate that $\vec{b}_0$ and $A_0$ are key to the convergence and the operations in Eq.~\ref{eq:b_trans}. This observation lays a foundation for the simulation study in Section~\ref{sec:sim}, where we study the effect of the ambiguity levels of $A_0$ and $\vec{b}_0$ in generalized settings.

\begin{table*}[t]
\scriptsize
\begin{center}
\resizebox{\textwidth}{!}{%
\begin{tabular}{l l l l l l}
\toprule
\textbf{Task} & \textbf{Datasets} &\textbf{\textsc{baseline}} &\textbf{Cls. Type}  & \textbf{\# Classes}    &\train{} / \val{} / \test{}\\ 
\midrule

\multirow{2}{*}{Entity Typing } 
& \multirow{2}{*}{\shortstack[l]{Ultrafine Entity Typing\\~\cite{choi-etal-2018-ultra}}}
  & Multitask~\cite{choi-etal-2018-ultra}  
  &\multirow{2}{*}{Multi-label}  &\multirow{2}{*}{10331}	
  &\multirow{2}{*}{1998 / 1998 / 1998} \\
  & &Denoised~\cite{onoe-durrett-2019-learning} & & & \\
\midrule

\multirow{2}{*}{\shortstack[l]{Relation\\Extraction}}
&\multirow{2}{*}{\shortstack[l]{DialogRE\\~\cite{yu-etal-2020-dialogue}}}
& BERT~\cite{yu-etal-2020-dialogue}
&\multirow{2}{*}{Multi-label} &\multirow{2}{*}{36} &\multirow{2}{*}{5997 / 1914 / 1862} \\
& &BERTs~\cite{yu-etal-2020-dialogue} & & &
\\
\midrule

\shortstack[l]{Image\\Classification}
 &\shortstack[l]{ImageNet\\~\cite{imagenet_cvpr09}} &\shortstack[l]{ResNet-50\\~\cite{he2016deep}}  &Single-label & 1000 & 1,281,167 / 50,000 / - \\
\bottomrule
\end{tabular}
}
\caption{Evaluation tasks and off-the-shelf \baseline{} methods used. Multiple \baseline{} architectures types are evaluated in our experiments.}
\label{tab:datasets}
\end{center}
\end{table*}

\begin{table*}
\begin{center}
\resizebox{\textwidth}{!}{%
\begin{tabular}{l l l l l l }
\toprule
\textbf{Task} & \textbf{Datasets}  & \textbf{\baseline}  & \textbf{$\alpha$}    &\textbf{$d$} &\textbf{$\tau$} \\ 
\midrule

\multirow{2}{*}{Entity Typing} 
& \multirow{2}{*}{\shortstack[l]{Ultrafine\\ Entity Typing} }
  &Multitask  &16	& 1  & 0.25  \\
& &Denoised   &22   & 2  & 0.75  \\
\midrule

\multirow{2}{*}{\shortstack[l]{Relation\\Extraction}}
& \multirow{2}{*}{DialogRE} & BERT &5, 0.7, 8, 5, 0.6 	&2, 3, 1, 3, 2   &0.75, 0.25, 0.25, 0.75, 0.5  \\
&&BERTS  &0.8, 7, 4, 7, 4 	&5, 1, 1, 1, 1    &0.25, 0.5, 0.5, 0.75, 0.25  \\

\midrule
\multirow{6}{*}{\shortstack[l]{Image \\Classification}}
 &ImageNet-Original   &\multirow{6}{*}{ResNet50} &0.6	&1  & \multirow{6}{*}{0.75}  \\
 &ImageNet-Blur ($\sigma = 2$)  &   &0.7	&2  &   \\
 &ImageNet-Blur ($\sigma = 4$)  &   &0.9	&5  &  \\
 &ImageNet-Blur ($\sigma = 8$)  &   &0.9	&3  &   \\
 &ImageNet-Blur ($\sigma = 16$) &   &0.5	&1  &   \\
 &ImageNet-Blur ($\sigma = 32$) &   &0.6	& 1  &   \\
\bottomrule
\end{tabular}
} 
\caption{Hyper-parameters used for all experiment. We report details of all the output from five runs of DialogRE dataset.}
\label{suptab:hp}
\end{center}
\end{table*}

\begin{table*}[!t]
\begin{center}
\resizebox{0.8\textwidth}{!}{%
\begin{tabular}{l l l c c }
\toprule
\textbf{Task} & \textbf{Datasets}  
& \textbf{\baseline}  & \textbf{\val}    &\textbf{\test} \\ 
\midrule

\multirow{2}{*}{Entity Typing} &\multirow{2}{*}{\shortstack[l]{Ultrafine\\ Entity Typing} }
  &Multitask  &14269 $\vert$ 0.07\%   &19007 $\vert$ 0.09\% \\
& &Denoised   &\phantom{0}4454 $\vert$ 0.02\%   &\phantom{0}4601 $\vert$ 0.02\% \\
\midrule

\multirow{11}{*}{\shortstack[l]{Relation\\Extraction}}
& \multirow{10}{*}{DialogRE} & \multirow{5}{*}{BERT}
  &\phantom{0}1478 $\vert$ 2.15\%  & \phantom{0}1430 $\vert$ 2.13\% \\
&&&\phantom{0}4598 $\vert$ 6.67\%  & \phantom{0}4512 $\vert$ 6.73\% \\
&&&\phantom{0}4126 $\vert$ 5.99\% & \phantom{0}3993 $\vert$ 5.96\% \\
&&&\phantom{0}1210 $\vert$ 1.76\% &\phantom{0}1148 $\vert$ 1.71\% \\
&&&\phantom{0}2610 $\vert$ 3.79\% &\phantom{0}1148 $\vert$ 1.71\% \\
\cmidrule{3-5}

&&\multirow{5}{*}{BERTs} 
&\phantom{0}3725 $\vert$ 5.41\%	&\phantom{0}3841 $\vert$ 5.73\%\\
&&&\phantom{0}1539 $\vert$ 2.23\%	&\phantom{0}1513 $\vert$ 2.26\%\\
&&&\phantom{0}1579 $\vert$ 2.29\%	&\phantom{0}1547 $\vert$ 2.31\%\\
&&&\phantom{00}885 $\vert$ 1.28\%	&\phantom{00}859 $\vert$ 1.28\%\\
&&&\phantom{0}3131 $\vert$ 4.54\%	&\phantom{0}3153 $\vert$ 4.70\%\\

\midrule

\multirow{6}{*}{\shortstack[l]{Image \\Classification}}
 &ImageNet-Original   &\multirow{6}{*}{ResNet50} &13484 $\vert$ 26.97\%  &\multirow{6}{*}{-}   \\
 &ImageNet-Blur ($\sigma = 2$)  &   &24788 $\vert$ 49.58\% &    \\
 &ImageNet-Blur ($\sigma = 4$)  &   &34421 $\vert$ 68.84\% &   \\
 &ImageNet-Blur ($\sigma = 8$)  &   &41230 $\vert$ 82.46\% & \\
 &ImageNet-Blur ($\sigma = 16$) &   &45721 $\vert$ 91.44\% &     \\
 &ImageNet-Blur ($\sigma = 32$) &   &48889 $\vert$ 97.78\%  &   \\
\bottomrule
\end{tabular}
}
\caption{Number of ambiguous examples in \val{} and \test{} sets (absolute $\vert$ relative ($\%$)).}
\label{suptab:amb_nums}
\end{center}\end{table*}

\section{Reproducibility Details}
\label{supsec:detail}
The experiments in this work do not require training or GPU.
We either download the publicly available pre-trained model checkpoints, or obtain the model output on \val~and \test~sets
from the researchers who propose the \baseline{}s.

\subsection{Datasets and \textsc{Baseline}s}
\label{subsec:expsetup}

The statistics of the evaluated tasks and the associated datasets are shown in Table~\ref{tab:datasets}. All evaluation and optimization protocols are based on the practices of the corresponding \baseline{}s for each task.

\paragraph{Ultrafine Entity Typing}
This task is to predict a set of semantic types of a given entity mention within a sentence. 
The dataset~\cite{choi-etal-2018-ultra} contains 10,331 entity types, including coarse, fine and ultra-fine grained classes.
Since each entity can have more than one types, this task is a multi-label classification problem.
Following previous practices~\cite{ling2012fine, choi-etal-2018-ultra}, we adopt the loose Macro and loose Micro F1 score as metrics.
We optimize the loose Macro F1 score on the \val{} set to select hyperparameters.
Two existing classifiers are evaluated: (1) Multitask~\cite{choi-etal-2018-ultra}: a model proposed along with the dataset and utilizes a LSTM-based AttentiveNER model~\cite{shimaoka-etal-2016-attentive}; (2) Denoised~\cite{onoe-durrett-2019-learning}: uses denoised distant training data.

\paragraph{DialogRE}
The task of Dialogue-based relation extraction (DialogRE)~\cite{yu-etal-2020-dialogue} dataset is to predict one or more types of relations between two entities mentioned in dialogues.
We applied \can to the two baselines proposed in the paper: BERT, and BERTs. 
Each experiment has five runs following the practices of~\citet{yu-etal-2020-dialogue}, so we report the average Macro F1 and Micro F1. The Micro F1 is used for the parameter optimization following~\citet{yu-etal-2020-dialogue}.

\subsection{Hyperparameters}
A hold-out validation set is required by our method. In practice, this can be the same set for fine tuning other hyperparameters during training neural classifiers.
Our assumption is that the different splits of a dataset (\train, \val, \test) are sampled from the same distribution.
The training class distribution is utilized as prior $\Lambda_q$.
We use the original \val{} set of the datasets to optimize for $\alpha \in \{0.1, \dots, 0.9, 1, \dots, 35 \}$, iteration number $d \in \{1, \dots, 5\}$, and the ambiguous threshold $\tau \in \{0.25, 0.5, 0.75\}$. Table~\ref{suptab:hp} summarizes all the hyperparameters used in this paper.
Table~\ref{suptab:amb_nums} presents resulting number of ambiguous distributions to re-adjust. 
We find that the optimized values of $d$ are mostly $\leq 2$ (77.3 \% of all experiments), which is in line with the discussions previously in the main text.

\end{document}